# Building a Silver-Standard Dataset from NICE Guidelines for Clinical LLMs

Qing DING[1], Eric H. Q. ZHANG[2], Felix JOZSA[3], and Julia IVE[4]
*University College London*
ORCid ID: Julia Ive https://orcid.org/0000-0002-3931-3392

**Abstract.** Large language models (LLMs) are increasingly used in healthcare, yet standardised benchmarks for evaluating guideline-based clinical reasoning are missing. This study introduces a validated dataset derived from publicly available guidelines across multiple diagnoses. The dataset was created with the help of GPT and contains realistic patient scenarios, as well as clinical questions. We benchmark a range of recent popular LLMs to showcase the validity of our dataset. The framework supports systematic evaluation of LLMs' clinical utility and guideline adherence. The dataset is publicly available at https://github.com/julia-ive/guidelines_qa.

## 1. Introduction

Clinical guidelines are essential for standardised patient care, offering evidence-based instructions for diagnosis and treatment [2]. While AI, particularly large language models (LLMs), shows promise in automating guideline adherence, these models can hallucinate or produce errors, posing risks in clinical settings. Most current LLM-based solutions rely on proprietary or institution-specific guidelines, limiting transparency and reproducibility [3,4]. To ensure safe and effective integration, publicly accessible benchmarks are needed to evaluate model reliability and alignment with clinical standards. Those currently do not exist.

To address these challenges, this study introduces a validated, publicly accessible dataset derived from clinical guidelines published by the National Institute for Health and Care Excellence (NICE, https://www.nice.org.uk/guidance), the UK's official body for evidence-based health recommendations. Covering multiple specialties, the dataset includes actionable clinical statements created using state-of-the-art knowledge distillation methods via optimised prompting and LLM-based filtering [5,6], paired with realistic patient scenarios and clinical questions. The methodology ensures clinical relevance through expert and LLM-based review. As a use case, several popular models such as GPT, Qwen and DeepSeek were benchmarked on the dataset, with outputs evaluated against reference statements using both automated metrics and manual assessment. By grounding the evaluation in realistic patient scenarios and clinically actionable statements derived from authoritative NICE guidelines, the framework ensures that model performance is tested against situations that closely resemble actual decision-making in healthcare.

Corresponding Author: Julia Ive, j.ive@ucl.ac.uk

## 2. Methodology

*2.1. Initial data collection and preprocessing*

Ten NICE clinical guidelines were selected under the guidance of clinical experts in the team to ensure relevance to general practice (ovarian cancer, gastro-oesophageal reflux disease, diabetes (types 1&2), heart failure, UTI, hypertension, thyroid disease, chronic kidney disease, anaphylaxis) [8-17]. We extracted clinical recommendations from NICE guideline PDFs via structural analysis of the contents or, when necessary, prompted manual input of page ranges. We then cleaned and segmented the extracted text into coherent paragraphs.

*2.2. Model Selection and Generation Process*

This study uses OpenAI's o4-mini model via its official API (temperature set to 0) to extract and generate clinical data. The selection of o4-mini is primarily based on its context processing capabilities and cost. To ensure robustness and clinical relevance, a two-step strategy was adopted to create the dataset.

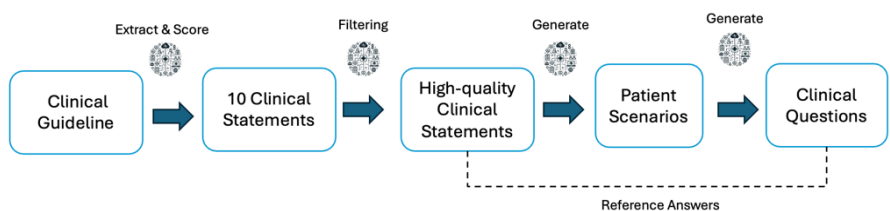

**Figure 1.** NICE guideline to clinical scenario generation workflow

Figure 1 illustrates workflow for generating clinically relevant scenario–question pairs from NICE guidelines: (1) Extract & score: clinical guideline text is processed to extract 10 candidate clinical statements. (2) Filtering: statements are evaluated and filtered to retain high-quality, actionable recommendations. (3) Generate patient scenarios: each validated statement is used to create realistic patient scenarios. (4) Generate clinical questions: clinical questions are derived from scenarios, with original statements serving as reference answers.

For each guideline, the o4-mini LLM was used to identify 10 clinically actionable statements based on a structured prompt (see Figure Prompt 1 in the Appendix). Each statement was required to meet such criteria as consistency with the original guideline text, precise reference locations within the guideline, specific, implementable guidance, including clear diagnostic criteria, defined treatment dosages, explicit decision rules. General principles or vague recommendations were excluded.

In the second step, each extracted statement was evaluated using an intelligent scoring system (Prompt 2 in the Appendix), developed under clinical oversight. The evaluation framework consisted of five binary-scored dimensions: clinical presentation

Corresponding Author: Julia Ive, j.ive@ucl.ac.uk

information, drug treatment guidance, diagnosis and decision-making, risk management and monitoring, as well as population/individualised adjustment.

Each dimension was scored as yes (1) or no (0), with a maximum score of 5. Statements scoring zero were excluded to ensure clinical relevance. The minimum threshold for inclusion was set at one point, based on iterative prompt refinement and expert review. In the final step, each validated statement was used to generate five distinct patient scenario–question pairs, resulting in approximately 500 scenario–question–answer triplets. The original guideline statements served as reference answers. All prompt designs were iteratively improved under the clinician oversight, and model outputs were manually reviewed by a clinical expert to ensure accuracy, consistency, and reliability.

We evaluated the created dataset using automatic metrics, as well as human expert (FJ) manual evaluation. In the evaluation of clinical statements, the expert used the same assessment criteria as o4-mini to validate the generated results (150 randomly selected triplets) across five core dimensions, thereby ensuring their effectiveness in terms of clinical usability.

Extrinsic evaluation measures dataset quality indirectly from a functional perspective by testing model performance in downstream tasks. We selected the following mainstream models: Qwen3-235b-a22b-instruct-2507, OpenAI GPT-5, Phi4-mini-instruct, and DeepSeek-R1. We used the default settings from the Replicate API to run those models (https://replicate.com) with Prompt 4 in Appendix 1.

## 3. Results

### 3.1. Dataset description & intrinsic evaluation

The statistics over the generated dataset are provided in Table 1 below. Initially, we extracted 522 clinical statements, each paired with a realistic patient scenario and corresponding clinical question, derived from multiple NICE guideline sections. Twenty-two statements were excluded after scoring zero in the o4-mini evaluation using the final prompt, indicating lack of clinical relevance (see examples of statements in Appendix 2). Our clinical expert reviewed 150 randomly selected samples, confirming that the scenarios are realistic and generated questions consistently reflected the core medical content of the reference answers, supporting the dataset's validity for benchmarking LLMs in clinical reasoning.

**Table 1. NICE guideline generated scenario dataset statistics**

|  | Average | Standard Deviation | Maximum Words | Median Words |
|---|---|---|---|---|
| Statement | 27.96 | 13.20 | 87 | 26.0 |
| Question | 12.75 | 3.17 | 26 | 12.0 |
| Scenario | 31.10 | 7.77 | 60 | 30.0 |

Corresponding Author: Julia Ive, j.ive@ucl.ac.uk

*3.2. Extrinsic evaluation*

We employed the following automated metrics:
- ROUGE: N-gram overlap between generated and reference responses [7]
- BLEU: precision against reference responses [18]
- METEOR: F1 measure on word synonym and stem [1]
- BERTScore-F1: token similarity between candidate sentence and reference sentence using contextual embeddings [20]

As shown in Appendix 3, automated evaluation results across leading large language models demonstrate significant performance variations, with Qwen3-235b-a22b-instruct-2507 establishing clear dominance across all metrics. The model achieved exceptional scores in ROUGE-1 (0.6222), ROUGE-2 (0.5598), ROUGE-L (0.5674), BLEU (0.3031), METEOR (0.6834), and BERTScore-F1 (0.9227), setting the benchmark for high-quality response generation and reference alignment.

OpenAI GPT-5 demonstrated competitive but notably lower performance, achieving ROUGE-1 (0.5447), ROUGE-2 (0.5003), ROUGE-L (0.5086), BLEU (0.2242), METEOR (0.6482), and BERTScore-F1 (0.9139). Despite maintaining strong semantic coherence as indicated by its high BERTScore-F1, GPT-5 showed substantial gaps compared to Qwen3, particularly in lexical overlap metrics with 12.5% lower ROUGE-1 and 10.6% lower ROUGE-2 scores.

Phi4-mini-instruct occupied the middle tier, delivering moderate performance across ROUGE-1 (0.5199), ROUGE-2 (0.4237), ROUGE-L (0.4759), BLEU (0.2688), METEOR (0.5584), and BERTScore-F1 (0.9068). The model's performance reflected significant deficits of 16.4% in ROUGE-1 and 24.3% in ROUGE-2 compared to the leading Qwen3 model, though it maintained reasonable semantic similarity as evidenced by its BERTScore-F1.

DeepSeek-R1 recorded the lowest scores across all metrics: ROUGE-1 (0.4278), ROUGE-2 (0.3820), ROUGE-L (0.4002), BLEU (0.2015), METEOR (0.5292), and BERTScore-F1 (0.8837), performing 31.2% lower on ROUGE-1 and 31.8% lower on ROUGE-2 compared to Qwen3. This performance difference appears to be related to DeepSeek-R1's extended reasoning approach, where procedures may inadvertently compromise evaluation performance when models prioritise internal reasoning over direct instruction following, resulting in outputs that were semantically different from the evaluation targets despite potentially sound reasoning.

Manual inspection conducted by clinical expert on the two worst responses of the top two best performing models (Appendix 4 & 5) illustrates two types of clinical output errors to the simulated scenarios. The model either choses the wrong clinical information that does not map to the question and scenario (Qwen3 worst #1, GPT-5 #2), or the correct clinical information is identified but the text output is very verbose and contains large amounts of excess information (Qwen3 worst #2, GPT-5 worst #1).

Corresponding Author: Julia Ive, j.ive@ucl.ac.uk

## 4. Conclusion

This study presents the first publicly available benchmark for evaluating LLMs against clinical guidelines, using realistic patient scenarios and validated NICE-based recommendations. It includes comprehensive benchmarking of recent leading models such as GPT5, Qwen3, and DeepSeek-R1, providing a robust foundation for assessing guideline adherence and clinical reasoning performance.

## References

[1] Banerjee S, Lavie A. METEOR: An Automatic Metric for MT Evaluation with Improved Correlation with Human Judgments. In: Goldstein J, et al., editors. Proceedings of the ACL Workshop on Intrinsic and Extrinsic Evaluation Measures for Machine Translation and/or Summarization. Ann Arbor: Association for Computational Linguistics; 2005. p. 65-72.
[2] Guerra-Farfan E, et al. Clinical practice guidelines: The good, the bad, and the ugly. Injury. 2023;54 Suppl 3:S26-S29.
[3] Ive J, et al. Clean & clear: Feasibility of safe LLM clinical guidance [Internet]. arXiv; 2025 [cited 2025 Aug 23]. Available from: http://arxiv.org/abs/2503.20953.
[4] Kresevic S, et al. Optimization of hepatological clinical guidelines interpretation by large language models: a retrieval augmented generation-based framework. NPJ Digit Med. 2024;7(1):102.
[5] Kweon S, et al. Publicly shareable clinical large language model built on synthetic clinical notes. In: Findings of ACL 2024; 2024.
[6] Gu J, et al. A Survey on LLM-as-a-Judge [Internet]. arXiv; 2024 [cited 2025 Aug 23]. Available from: http://arxiv.org/abs/2411.15594.
[7] Lin CY. ROUGE: A Package for Automatic Evaluation of Summaries. In: Text Summarization Branches Out. Barcelona: Association for Computational Linguistics; 2004. p. 74-81.
[8] National Institute for Health and Care Excellence. Ovarian cancer: Recognition and initial management. Clinical guideline CG122. London: NICE; 2011. Available from: https://www.nice.org.uk/guidance/cg122.
[9] National Institute for Health and Care Excellence. Gastro-oesophageal reflux disease and dyspepsia in adults: Investigation and management. Clinical guideline CG184. London: NICE; 2014. Available from: https://www.nice.org.uk/guidance/cg184.
[10] National Institute for Health and Care Excellence. Type 1 diabetes in adults: Diagnosis and management. NICE guideline NG17. London: NICE; 2015. Available from: https://www.nice.org.uk/guidance/ng17.
[11] National Institute for Health and Care Excellence. Type 2 diabetes in adults: Management. NICE guideline NG28. London: NICE; 2015. Available from: https://www.nice.org.uk/guidance/ng28.
[12] National Institute for Health and Care Excellence. Chronic heart failure in adults: Diagnosis and management. NICE guideline NG106. London: NICE; 2018. Available from: https://www.nice.org.uk/guidance/ng106.
[13] National Institute for Health and Care Excellence. Urinary tract infection (lower): Antimicrobial prescribing. NICE guideline NG109. London: NICE; 2018. Available from: https://www.nice.org.uk/guidance/ng109.
[14] National Institute for Health and Care Excellence. Hypertension in adults: Diagnosis and management. NICE guideline NG136. London: NICE; 2019. Available from: https://www.nice.org.uk/guidance/ng136.
[15] National Institute for Health and Care Excellence. Thyroid disease: Assessment and management. NICE guideline NG145. London: NICE; 2019. Available from: https://www.nice.org.uk/guidance/ng145.
[16] National Institute for Health and Care Excellence. Chronic kidney disease: Assessment and management. NICE guideline NG203. London: NICE; 2021. Available from: https://www.nice.org.uk/guidance/ng203.
[17] National Institute for Health and Care Excellence. Use of AI in evidence generation: NICE position statement [Internet]. London: NICE; 2024. Available from: https://www.nice.org.uk/position-statements/use-of-ai-in-evidence-generation-nice-position-statement.
[18] Papineni K, et al. BLEU: a method for automatic evaluation of machine translation. In: Proceedings of the 40th Annual Meeting on Association for Computational Linguistics - ACL '02. Philadelphia: Association for Computational Linguistics; 2002. p. 311.
[19] Resuscitation Council UK. Emergency treatment of anaphylaxis: Guidelines for healthcare providers [Internet]. London: Resuscitation Council UK; 2021. Available from:

Corresponding Author: Julia Ive, j.ive@ucl.ac.uk

https://www.resus.org.uk/sites/default/files/2021-05/Emergency%20Treatment%20of%20Anaphylaxis%20May%202021_0.pdf

[20] Zhang T, Kishore V, Wu F, Weinberger KQ, Artzi Y. Bertscore: Evaluating text generation with bert. arXiv preprint arXiv:1904.09675. 2019.

Corresponding Author: Julia Ive, j.ive@ucl.ac.uk

# Appendix 1

**Prompt 1:**

"Extract the Ten most clinically actionable statements from the given guideline {guideline_text} to provide healthcare professionals with precise and implementable instructions for diagnosing and managing the condition. These extracted statements should be highly suitable for serving as direct, standalone answers to specific medical questions.

Before extracting statements, preprocess the guideline by removing all content related to research directions, social/ethical background, references, and author committee information, as these are not core parts of the guideline. However, retain clinically actionable content from appendices, including flowcharts, risk scores, diagnostic algorithms, and structured treatment pathways that provide clear clinical guidance.

Identify and extract statements from the guideline that meet the following criteria:

1) they must be identical to the original text of the guideline;
2) they must provide specific, step-by-step clinical guidance rather than broad recommendations. It should directly address ""how to,"" ""when to,"" ""what to do,"" or ""what to use"" in a clinical scenario. ;
3) they must include clear diagnostic criteria, treatment dosages, procedural steps, or structured decision-making rules that can be directly implemented by healthcare professionals;
4) they must avoid general principles, educational recommendations, or vague guidelines without direct clinical applicability.
5) If a sentence or bullet point includes multiple conditions, treatment options, or sequential clinical actions, split it into separate statements, each covering only one scenario or clinical step.

Use the original wording, but do not combine different steps or branches in one output.

When there are multiple actionable statements, return those that directly provide solutions, the next steps, or clear management guidance in their exact original wording and mark the point number (such as 1.2.1) at the end."

**Prompt 2:**

"You are a clinical guideline expert. Your task is to evaluate each of the following clinical statements based on the reference guideline provided.

Guideline Text:{guideline_text}
Statements to Evaluate:{statements}

Evaluation Criteria:
Q1. Does the statement describe the patient's symptoms, signs, clinical manifestations or chief complaints?
Q2. Does this statement involve drug treatment (such as initiating a new drug, adjusting the treatment plan or providing specific medication guidance)? Q3. Does this statement provide a positive choice of diagnostic methods, interpretation of diagnostic results, or describe a clear decision-making process based on findings?
Q4. Does this statement contain risk assessment, monitoring plans or specific clinical target values?
Q5. Does this statement provide adjustment suggestions for specific populations or clear treatment adjustment/termination criteria?

Output format:
statement 1:Q1=score; Q2=score;Q3=score;Q4=score;Q5=score；
Total=Q1+Q2+Q3+Q4+Q5
..."

Corresponding Author: Julia Ive, j.ive@ucl.ac.uk

**Prompt 3:**

"You are a clinical scenario generator assistant.
Based on the clinical guideline statement {statements}, generate:

1. Five realistic and concise general practitioner patient scenarios containing key patient clinical information.
2. A single-sentence question that leads to the statement as the correct answer for each scenario.
3. Demonstrate the relationship between the question and the scenario (e.g., For this patient) while avoiding repetition of previous clinical information from the scenario.
4. The statement is again used as the correct answer.

Output format:

### Scenario:
<The scenario you generated>

### Question:
<The question you generated>

### Answer:
<The original statement>"

**Prompt 4:**
[{
    "role": "system",
    "content": f"""You are a clinical guideline expert. Your task is to answer clinical questions by extracting the most relevant and actionable guidance from the provided medical guideline.

Medical Guideline:
{guideline_text}

Instructions for answering:
1. Extract the most clinically actionable statement(s) from the guideline that directly answer the question
2. Use the exact wording from the guideline - do not paraphrase or modify the text
3. Focus on specific, implementable clinical guidance that addresses "how to," "when to," "what to do," or "what to use"
4. Include clear diagnostic criteria, treatment dosages, procedural steps, or decision-making rules
5. If the guideline contains numbered sections (e.g., 1.2.1), include the reference number
6. Provide step-by-step guidance when the scenario requires sequential clinical actions
7. Keep the answer concise and directly relevant to the specific question asked
8. Avoid general principles or vague recommendations - provide precise, actionable instructions"""
  }, {
    "role": "user",
    "content": f"""Clinical Scenario:
{clinical_scenario}

Question:
{clinical_question}

Based on the guideline provided, extract and provide the exact statement(s) that answer this question. Use the original wording from the guideline and include any relevant reference numbers."""
  }]

Corresponding Author: Julia Ive, j.ive@ucl.ac.uk

**Appendix 2:**

| Guideline | Number of Statements | Example Statement |
|---|---|---|
| Urinary tract infection (lower) | 10 | Advise people with lower UTI about using paracetamol for pain, or if preferred and suitable ibuprofen. (3.1) |
| Type1 diabetes | 10 | Measure HbA$_{1c}$ levels every 3 to 6 months in adults with type 1 diabetes. [6.1] |
| Type2 diabetes | 10 | Offer standard-release metformin as firstline drug treatment to adults with type 2 diabetes. (7.3) |
| Thyroid disease | 10 | Before starting antithyroid drugs for adults, children and young people with hyperthyroidism, check full blood count and liver function tests. (6.21) |
| Ovarian cancer | 10 | If serum CA125 is 35 IU/ml or greater, arrange an ultrasound scan of the abdomen and pelvis. 1.2.2 |
| Hypertension in adults | 10 | If hypertension is not controlled in adults taking step 2 treatment, offer a combination of: an ACE inhibitor or ARB and a CCB and a thiazide-like diuretic. [2019] (4.45) |
| Gastro-oesophageal reflux disease and dyspepsia in adults | 10 | Offer people with GORD a full-dose PPI (see table 1 in appendix A) for 4 or 8 weeks. [6.2] |
| Diagnosis and management of heart failure | 10 | Introduce beta-blockers in a "start low, go slow" manner. [2010, amended 2018] (4.13) |
| Chronic kidney disease | 10 | Confirm an eGFR result of less than 60 ml/min/1.73 m2 in an adult not previously tested by repeating the test within 2 weeks. (1.8) |
| Anaphylaxis recommendation | 10 | Adult and child > 12 years: 500 micrograms IM (0.5 mL of 1 mg/mL adrenaline). (5.1.1) |

**Appendix 3: Automated Evaluation**

| Model | ROUGE 1 | ROUGE 2 | ROUGE L | BLEU | METEOR | BERTScore-F1 |
|---|---|---|---|---|---|---|
| Qwen3-235b-a22b-instruct-2507 | **0.6222** | **0.5598** | **0.5674** | **0.3031** | **0.6834** | **0.9227** |
| OpenAI GPT-5 | 0.5447 | 0.5003 | 0.5086 | 0.2242 | 0.6482 | 0.9139 |
| Phi4-mini-instruct | 0.5199 | 0.4237 | 0.4759 | 0.2688 | 0.5584 | 0.9068 |
| DeepSeek-R1 | 0.4278 | 0.3820 | 0.4002 | 0.2015 | 0.5292 | 0.8837 |

Corresponding Author: Julia Ive, j.ive@ucl.ac.uk

**Appendix 4: Qwen3-235b-a22b-instruct-2507 Top 2 Best & Worst Reponses**

| Performance | Clinical Scenario | Generated Answer | Reference Answer |
|---|---|---|---|
| Best response 1 | A 15-year-old adolescent male reports suprapubic discomfort and burning sensation on urination for 48 hours. He is afebrile, with a normal cardiovascular exam and clear lung fields. For this patient, what is the appropriate next step before starting antibiotics? | 1.9 Obtain a urine sample from children and young people with lower UTI before antibiotics are taken, and dipstick test or send for culture and susceptibility testing in line with the NICE guideline on urinary tract infection in under 16s. | 6. Obtain a urine sample from children and young people with lower UTI before antibiotics are taken, and dipstick test or send for culture and susceptibility testing in line with the NICE guideline on urinary tract infection in under 16s. (1.9) |
| Best response 2 | A 63-year-old woman with refractory ascites despite diuretics is referred by her GP; pelvic ultrasound identifies a complex ovarian mass and CT confirms peritoneal carcinomatosis, with CA-125 elevated to 1,050 U/mL. For this patient, what key diagnostic step should be completed before starting cytotoxic chemotherapy? | 1.2.4.1 If offering cytotoxic chemotherapy to women with suspected advanced ovarian cancer, first obtain a confirmed tissue diagnosis by histology (or by cytology if histology is not appropriate) in all but exceptional cases. | 3. If offering cytotoxic chemotherapy to women with suspected advanced ovarian cancer, first obtain a confirmed tissue diagnosis by histology (or by cytology if histology is not appropriate) in all but exceptional cases. 1.2.4.1 |
| Worst response 1 | A 48-year-old woman of African Caribbean descent with controlled type 2 diabetes is found to have clinic readings of 150/95 mmHg on two occasions despite lifestyle advice. Her ECG and renal profile are unremarkable. For this patient, which guideline recommendation should guide further assessment? | "1.4.32 Offer an ACE inhibitor or an ARB to adults starting step 1 antihypertensive treatment who:<br>• have type 2 diabetes and are of any age or family origin (see also recommendation 1.4.30 for adults of Black African or African–Caribbean family origin) or<br>• are aged under 55 but not of Black African or African–Caribbean family origin. [2019]<br><br>1.4.30 When choosing antihypertensive drug treatment for adults of Black African or African–Caribbean family origin, consider an angiotensin II receptor blocker (ARB), in preference to an angiotensin-converting enzyme (ACE) inhibitor. [2019]" | 6. Confirm diagnosis of hypertension in people with a clinic blood pressure of 140/90 mmHg or higher and ABPM daytime average or HBPM average of 135/85 mmHg or higher. [2.8] |
| Worst response 2 | A 65-year-old asthmatic on beta-blockers presents with life-threatening | "For this patient, the next step in epinephrine administration is: | 10. Mix 1 mg (1 mL of 1 mg/mL [1:1 000]) adrenaline in 100 mL of 0.9% sodium |

Corresponding Author: Julia Ive, j.ive@ucl.ac.uk

| | bronchospasm, poor air entry, worsening hypoxia, and hypotension despite repeated inhaled bronchodilators and IM epinephrine. For this patient, what is the next step in epinephrine administration? | ""Patients with no improvement in respiratory and/or cardiovascular symptoms due to anaphylaxis, despite two appropriate doses of adrenaline, should be started on an adrenaline infusion."" (6.6)<br><br>""Seek urgent expert help to establish a low-dose, IV adrenaline infusion. IV adrenaline should be given only by experienced specialists in an appropriate setting."" (6.6)<br><br>""Continue to repeat IM adrenaline after 5 minutes until the infusion has been started. If an IV infusion cannot be administered safely (e.g. due to a patient being outside a hospital setting), continue to repeat IM adrenaline after every 5 minutes while life-threatening respiratory or cardiovascular features persist, and seek urgent ambulance support/expert help."" (6.6)" | chloride and connect using an infusion pump via a dedicated line. 6.6 |
|---|---|---|---|

## Appendix 5: OpenAI Gpt-5

| Performance | Clinical Scenario | Generated Answer | Reference Answer |
|---|---|---|---|
| Best response 1 | A 17-year-old girl suffers from chronic headaches and uses over-the-counter naproxen daily for the last 10 months. She presents for her routine health check. For this patient, what annual test should be performed? | "Monitor GFR at least annually in adults, children and young people who are taking medicines that can adversely affect kidney function, such as calcineurin inhibitors (for example, ciclosporin or tacrolimus), lithium or non-steroidal anti-inflammatory drugs (long-term chronic use of NSAIDs)." [1.1.20] | 7. Monitor GFR at least annually in adults, children and young people who are taking medicines that can adversely affect kidney function, such as calcineurin inhibitors (for example, ciclosporin or tacrolimus), lithium or non-steroidal anti-inflammatory drugs (long-term chronic use of NSAIDs). (1.20) |
| Best response 2 | A 35-year-old woman with newly diagnosed Hashimoto's primary hypothyroidism started levothyroxine 50 mcg daily two months ago; her first TSH at four weeks was 5.8 mU/L. For this patient, what is the recommended | 4.3 For adults who are taking levothyroxine for primary hypothyroidism, consider measuring TSH every 3 months until the level has stabilised (2 similar measurements within the | 8. For adults who are taking levothyroxine for primary hypothyroidism, consider measuring TSH every 3 months until the level has stabilised (2 similar measurements within the reference range 3 months |

Corresponding Author: Julia Ive, j.ive@ucl.ac.uk

| | | | |
|---|---|---|---|
| | schedule for monitoring TSH until her levels stabilise and thereafter? | reference range 3 months apart), and then once a year. | apart), and then once a year. (4.3) |
| Worst response 1 | A 65-year-old asthmatic on beta-blockers presents with life-threatening bronchospasm, poor air entry, worsening hypoxia, and hypotension despite repeated inhaled bronchodilators and IM epinephrine.<br><br>For this patient, what is the next step in epinephrine administration? | "• "If features of anaphylaxis persist despite two doses of IM adrenaline, follow the refractory anaphylaxis algorithm (see section 6) and call for expert support to allow an intravenous adrenaline infusion to be started." (5.1 Adrenaline [Epinephrine]; 5.1)<br><br>• "Patients with no improvement in respiratory and/or cardiovascular symptoms due to anaphylaxis, despite two appropriate doses of adrenaline, should be started on an adrenaline infusion." (6.6 IV adrenaline infusion)<br><br>• "Seek urgent expert help to establish a low-dose, IV adrenaline infusion. IV adrenaline should be given only by experienced specialists in an appropriate setting." (6.6 IV adrenaline infusion)<br><br>• "Continue to repeat IM adrenaline after 5 minutes until the infusion has been started. If an IV infusion cannot be administered safely (e.g. due to a patient being outside a hospital setting), continue to repeat IM adrenaline after every 5 minutes while life-threatening respiratory or cardiovascular features persist, and seek urgent ambulance support/expert help." (6.6 IV adrenaline infusion)<br><br>• "Adrenaline may be less effective in patients treated with a beta blocker. In these cases, consider giving glucagon when symptoms remain refractory to adrenaline infusion and adequate fluid resuscitation." (6.7.4 Other medications)" | 10. Mix 1 mg (1 mL of 1 mg/mL [1:1 000]) adrenaline in 100 mL of 0.9% sodium chloride and connect using an infusion pump via a dedicated line. 6.6 |

Corresponding Author: Julia Ive, j.ive@ucl.ac.uk

| | | | |
|---|---|---|---|
| Worst response 2 | A 55-year-old woman with type 2 diabetes on multiple daily injections of insulin glulisine (Apidra) reports cost-related nonadherence and recent glycemic variability. For this patient, what conversation is most appropriate regarding her insulin therapy? | "- "For adults with type 2 diabetes starting insulin therapy, provide a structured programme using active insulin dose titration that encompasses: • injection technique, including rotating injection sites and avoiding repeated injections at the same point within sites • continuing telephone support • self-monitoring • dose titration to target levels • dietary advice • the DVLA's Assessing fitness to drive: a guide for medical professionals • managing hypoglycaemia • managing acute changes in plasma glucose control • support from an appropriately trained and experienced healthcare professional." [1.7.24]<br><br>- "Start insulin therapy for adults with type 2 diabetes from a choice of the following insulin types and regimens: • Offer neutral protamine Hagedorn (NPH) insulin injected once or twice daily according to need. • Consider starting both NPH and short-acting insulin (particularly if the person's HbA1c is 75 mmol/mol [9.0%] or higher), administered either: ─ separately or ─ as a pre-mixed (biphasic) human insulin preparation. • Consider, as an alternative to NPH insulin, using insulin detemir or insulin glargine if: ─ the person needs help from a carer or healthcare professional to inject insulin, and use of insulin detemir or insulin glargine would reduce the frequency of injections from twice to once daily or ─ the person's lifestyle is restricted by recurrent symptomatic hypoglycaemic episodes or ─ the person would otherwise need twice-daily NPH insulin injections in combination with oral glucose-lowering drugs. • Consider pre-mixed | 10. When people are already using an insulin for which a lower cost biosimilar is available, discuss the possibility of switching to the biosimilar. [2021] (7.32) |

Corresponding Author: Julia Ive, j.ive@ucl.ac.uk

| | | | |
|---|---|---|---|
| | | (biphasic) preparations that include short-acting insulin analogues, rather than pre-mixed (biphasic) preparations that include short-acting human insulin preparations, if: ─ the person prefers injecting insulin immediately before a meal or ─ hypoglycaemia is a problem or ─ blood glucose levels rise markedly after meals." [1.7.26]<br><br>- "When starting an insulin for which a biosimilar is available, use the product with the lowest acquisition cost." [1.7.30]<br><br>- "Ensure the risk of medication errors with insulins is minimised by following the Medicines and Healthcare products Regulatory Agency (MHRA) guidance on minimising the risk of medication error with high strength, fixed combination and biosimilar insulin products, which includes advice for healthcare professionals when starting treatment with a biosimilar." [1.7.31]<br><br>- "When people are already using an insulin for which a lower cost biosimilar is available, discuss the possibility of switching to the biosimilar. Make a shared decision with the person after discussing their preferences." [1.7.32]" | |

Corresponding Author: Julia Ive, j.ive@ucl.ac.uk